\pdfoutput=1

\documentclass[11pt]{article}

\usepackage[]{ACL2023}

\usepackage{times}
\usepackage{latexsym}

\usepackage[T1]{fontenc}

\usepackage[utf8]{inputenc}

\usepackage{microtype}

\usepackage{inconsolata}
\usepackage{tabularx}
\usepackage{booktabs}
\usepackage{amsmath}
\usepackage{pifont}
\usepackage{xcolor}
\definecolor{green1}{RGB}{34,139,34}
\definecolor{red1}{RGB}{139,0,0}
\usepackage{enumerate}
\usepackage{multirow}
\usepackage{tablefootnote}
\usepackage{tikz}
\usepackage{amssymb}

\newcommand{\cmark}{\textcolor{green1}{\ding{51}}}%
\newcommand{\xmark}{\textcolor{red1}{\ding{55}}}%

\title{CQE: A Comprehensive Quantity Extractor}

\author{Satya Almasian\thanks{~~These authors contributed equally to this work.}, Vivian Kazakova$^{*}$, Philip Göldner \and Michael Gertz \\
  Institute of Computer Science, Heidelberg University \\
  \texttt{\{almasian,gertz\}@informatik.uni-heidelberg.de}\\
   \texttt{\{vivian.kazakova,goeldner\}@stud.uni-heidelberg.de}}

\begin{document}
\maketitle
\begin{abstract}
  	Quantities are essential in documents to describe factual information.
 They are ubiquitous in application domains such as finance, business,
 medicine, and science in general. Compared to other information
 extraction approaches, interestingly only a few works exist that describe
 methods for a proper extraction and representation of quantities in
 text.\\
 \noindent In this paper, we present such a comprehensive
 quantity extraction framework from text data. It efficiently detects
 combinations of \emph{values} and \emph{units}, the behavior of a
 quantity (e.g., rising or falling), and the \emph{concept} a quantity
 is associated with. Our framework makes use of dependency parsing and
 a dictionary of units, and it provides for a proper normalization and
 standardization of detected quantities. Using a novel dataset for
 evaluation, we show that our open source framework outperforms other
 systems and -- to the best of our knowledge -- is the first to detect
 concepts associated with identified quantities. The code and data
 underlying our framework are available at
 \url{https://github.com/vivkaz/CQE}.


\end{abstract}

\section{Introduction}
\label{sec:intro}
Quantities are the main tool for conveying factual and accurate information. News articles are filled with social and financial trends, and technical documents use measurable values to report their findings.  Despite their significance, a comprehensive system for quantity extraction and an evaluation framework to compare the performance of such systems is not yet at hand.  In the literature, a few works directly study quantity extraction, but their focus is limited to physical and science domains~\cite{DBLP:conf/doceng/FoppianoRIT19}.  Quantity extraction is often part of a larger system, where identification of quantities is required to improve numerical understanding in retrieval or textual entailment tasks~\cite{roy-etal-2015-reasoning,DBLP:conf/wsdm/LiFLLZ21,DBLP:conf/kdd/SarawagiC14,DBLP:conf/sigir/BanerjeeCR09,DBLP:conf/sigir/MaiyaVW15}.  Consequently, their performance is measured based on the downstream task, and the quality of the extractor, despite its contribution to the final result, is not separately evaluated.  Therefore, when in need of a quantity extractor, one has to resort to a number of open source packages, without a benchmark or a performance guarantee.
\noindent Since quantity extraction is rarely the main objective, the capabilities of the available systems and their definition of quantity vary based on the downstream task.  As a result, the context information about a quantity is reduced to the essentials of each system.  Most systems consider a quantity as a number with a measurable and metric \emph{unit}~\cite{DBLP:conf/doceng/FoppianoRIT19}.  However, outside of scientific domains any noun phrase describing a \emph{value} is a potential \emph{unit}, e.g., ``5 bananas''.  Moreover, a more meaningful representation of quantities should include their behaviour and associated \emph{concepts}.  For example, in the sentence ``DAX fell 2\% and S\&P gained more than 2\%'', the \emph{value/unit} pair $\langle$\emph{2, percentage}$\rangle$ indicates two different quantities in association with different concepts, DAX and S\&P, with opposite behaviours, \emph{decreasing} and \emph{increasing}, subtleties not captured by simplified models.  \noindent In this paper, we present a comprehensive quantity extraction (CQE) framework.  Our system is capable of extracting standardized \emph{values}, physical and non-physical \emph{units}, \emph{changes} or trends in the values and \emph{concepts} associated with detected values.  Furthermore, we introduce NewsQuant, a new benchmark dataset for quantity extraction, carefully selected from a diverse set of news articles in the categories of economics, sports, technology, cars, science, and companies.  Our system outperforms other libraries and extends on their capabilities to extract \emph{concepts} associated with values.  Our software and data are publicly available. By introducing a strong baseline and a novel dataset, we aim to motivate further research in this field.

\section{Related Work}
\label{sec:related}
In literature, quantity extraction is mainly a component of a larger system for textual entailment or search. 
The only work that solely focuses on quantity extraction is Grobid-quantities~\cite{DBLP:conf/doceng/FoppianoRIT19}, which uses three Conditional Random Field models in a cascade to find \emph{value/unit} pairs and to determine their relation, where the \emph{units} are limited to the scientific domain, a.k.a.~\emph{SI units}.

\noindent Roy et al.'s~\cite{roy-etal-2015-reasoning} definition of a quantity is closer to ours and is based on Forbus' theory~\cite{DBLP:journals/ai/Forbus84}. A quantity is a  (\emph{value}, \emph{unit}, \emph{change}) triplet, and noun-based units are also considered. 
Extraction is performed as a step in their pipeline for quantity reasoning in terms of textual entailment.  
Although they only evaluate on textual entailment, the extractor is released as part of the CogComp natural language processing libraries, under the name Illinois Quantifier\footnote{\url{https://github.com/CogComp/cogcomp-nlp/tree/master/quantifier} Last accessed: April 15, 2023}. 

\noindent Two prominent open source libraries for quantity extraction are (a) Recognizers-Text~\cite{soft:recognizers-text,chen2023dataset} from Microsoft and (b) Quantulum3~\footnote{\url{https://github.com/nielstron/quantulum3} Last accessed: April 15, 2023}. Recognizers-Text uses regular expressions for the resolution of numerical and temporal entities in 10 languages. The system has separate models for the extraction of \emph{value/unit} pairs for percentages, age, currencies, dimensions, and temperatures and is limited to only these quantity types.
Moreover, it cannot proactively distinguish the type of quantity for extraction and the user has to manually select the correct model. 
Quantulum3 uses regular expression to extract quantities and a dictionary of \emph{units} for normalization. For \emph{units} with similar surface forms, a classifier based on Glove embeddings~\cite{pennington-etal-2014-glove} is used for disambiguation, e.g.,
``pound'' as weight or currency.
Recognizers-Text is used in the work of \cite{DBLP:conf/wsdm/LiFLLZ21} to demonstrate quantity search, where the results are visualized in the form of tables or charts.  They define quantity facts as triplets of (\emph{related, value \& unit, time}).  \emph{Related} is the quantitative related information, close to our definition of \emph{concept}. 
However, it is not part of their quantity model but rather extracted separately using rules.
They utilize the quantity facts for visualization of results but do not evaluate their system or the quantity extraction module. 
QFinder~\cite{DBLP:conf/sigir/AlmasianB022} uses Quantulum3 in a similar way to demonstrate quantity search on news articles, but does not comment on the extractor's performance.

\noindent A number of other works utilize quantity extraction as part of their system. MQSearch~\cite{DBLP:conf/sigir/MaiyaVW15} extracts quantities with a set of regular expressions for a search engine on numerical information.  Qsearch~\cite{DBLP:conf/semweb/HoIPBW19} is another quantity search system, based on quantity facts extracted with the Illinois Quantifier. The works by \cite{DBLP:conf/sigir/BanerjeeCR09,DBLP:conf/kdd/SarawagiC14} focus on scoring quantity intervals in census data and tables.

\section{Extraction of Quantities}
In the following, we describe our quantity representation model and detail our extraction technique.

\subsection{Quantity Representation} In general, anything that has a count or is measurable is considered a quantity.  We extend upon the definition by \cite{roy-etal-2015-reasoning} to include concepts and represent a quantity by a tuple $\langle v,u,ch,cn \rangle$ with the following components:
\begin{enumerate}[1.]
\item \emph{Value ($v$):} A real number or a range of values, describing a magnitude, multitude, or duration, e.g., ``the car accelerates from 0 to 72 km/h'', has a range of $v=(0,72)$ and, ``the car accelerated to 72 km/h'' has a single value $v=72$.
\emph{Values} come in different magnitudes, often denoted by prefixes, and sometimes containing fractions, e.g., ``He earns 10k euros'' $\to$ $v=10000$, or ``1/5 th of his earnings''$\to$ $v=0.2$.
  
\item \emph{Unit ($u$):} A noun phrase defining the atomic unit of measure.  \emph{Units} are either part of a predefined set of known scientific and monetary types, or in a more general case, are noun phrases that refer to the multitude of an object, e.g., ``2 apples'' $\to$ $u=apple$.  The predefined set corresponds either to (a) \emph{scientific units} for measurement of physical attributes (e.g., ``2km'' has the \emph{scientific unit} ($u=kilometre$)), or (b) \emph{currencies}, as the unit of money (e.g., ``10k euros'' refers to a currency).  Predefined \emph{units} can have many textual or symbolic surface forms, e.g., ``euro'', ``EUR'', or ``\texteuro '', and their normalization is a daunting task.  Sometimes the surface forms coincide with other units, resulting in ambiguity that can only be resolved by knowing the context, e.g., ``She weighs 50 pounds'', is a measure of weight ($u=$pound-mass) and not a currency.

\item \emph{Change ($ch$):}  The modifier of the quantity  \emph{value}, describing how the \emph{value} is changing, e.g., ``roughly 35\$'' is describing an approximation. \cite{roy-etal-2015-reasoning} introduce four categories for \emph{change}:  $=$ (equal), $\sim$ (approximate), $>$ (more than), and $<$ (less than). These categories are mainly describing the bounds for a quantity.  We extend this definition by accounting for trends and add two more categories: $up$ and $down$ for increasing and decreasing trends, e.g., ``DAX fell 2\%'' indicates a downward trend ($ch=down$), while ``He weighs more than 50kg'' is indicating a bound ($ch=\text{`>`}$). 
\item \emph{Concept ($cn$):} \emph{Concepts} are either properties being measured or entities that the \emph{value} is referring to or is acting upon.  In the phrase ``DAX fell 2\%'' the quantity is measuring the worth of $cn=DAX$ or in ``The BMW Group is investing a total of \$200 million '' the investment is being made by $cn=BMW \ Group$. Sometimes a \emph{concept} is distributed in different parts of a sentence, e.g., ``The iPhone 11 has 64GB of storage. '' $\to cn=iPhone\ 11,storage$.
A \emph{concept} may or may not be present, e.g., ``200 people were at the concert'' has no concept.
\end{enumerate}

 \vspace{-0.2cm}
\subsection{Quantity Extraction}

Similar to previous work, we observed that quantities often follow a recurring pattern.  But instead of relying on regular expressions, we take advantage of linguistic properties and dependency parsing.  The input of our system is a sentence, and the output is a list of detected quantities.

\noindent
Example 1:
``In Europe, German DAX fell 0.4 pc, while the CAC40 in France gained 0.1.'' results in
\begin{itemize}
 
\item  \hspace{-0.1cm}$\langle v=0.4,\ u=percentage,$ \\
$ ch=down,\  cn=(German, DAX) \rangle$
\item \hspace{-0.2cm} $\langle v=0.1,\ u=percentage,$ \\
  $ ch=up,\ cn=(CAC40, France) \rangle$.
\end{itemize}
The extraction is performed in five stages, as described next.
\subsubsection{Pre-processing}
The pre-processing stage includes the removal of unnecessary punctuations, e.g., ``m.p.h'' $\to$ ``mph'', the addition of helper tokens, and other text cleaning steps.  
Numerals that do not fit the definition of a quantity, such as phone numbers and dates, are detected with regular expressions disregarded in further steps. 

\subsubsection{Tokenization} 
We perform a custom task-specific word tokenization. 
Our tokenizer is aware of separator patterns in \emph{values} and \emph{units} and avoids between-word splitting. 
For example, in the sentence ``A beetle goes from 0 to 80 km/h in 8 seconds.'', a normal tokenizer would split \emph{km/h} $\to$ \emph{(km, /, h)} but we will keep the \emph{unit} token intact. 
Another example is a numerical token containing punctuations, e.g., $2.33E\text{-}3$,  where naive tokenization changes the value.

\subsubsection{Value, Unit, and Change Detection}
The tokenized text is matched against a set of rules based on a dependency parsing tree and POS tags.  A set of 61 rules were created based on patterns observed and by studying previous work.  The comprehensive list of all rules is found in the repository of our project.  The rules are designed to find tokens associated with \emph{value}, \emph{unit}, and \emph{change}. 
\emph{Value/unit} pairs are often sets of numbers and nouns, numbers and symbols, or number and adjectives in various sentence structures. 
For ranges, the rules become more complex, as lower and upper bounds need to be identified using relational keywords such as ``from... to'' or ``between''.
\emph{Changes} are often adjectives or verbs that have a direct relation to a number and modify its value. 
Sometimes symbols before a number are also an indication of a \emph{change}, e.g., ``$\sim 10$'' describes an approximation.
In general, there are six \emph{change} categories.  $\sim$ for approximate equality, $=$ for exact equality, $>$ for greater than bounds, $<$ for less than bounds, $up$ denoting an increasing or upward trend, and $down$ for decreasing or downward trend. As an example of the extraction, we look at \emph{value}, \emph{unit} and \emph{change} detection for the two quantities in Example 1. 
Note that in this stage the surface forms are detected and not normalized values, e.g., ``pc'' versus ``percentage''.

\noindent
The NOUN$\textunderscore$NUM rule detects the surface form for the first \emph{value/unit} pair, (0.4, pc). Here, the \emph{value} has NUM as a POS-tag and is the immediate syntactic dependent of the \emph{unit} token, which is a noun or proper noun. 

\noindent
The LONELY$\textunderscore$NUM rule detects the \emph{value/unit} pair for the second quantity, namely (\emph{0.1, None}). If all other rules fail to find a \emph{value/unit} pair, this rule detects the number with the POS-tag NUM.

\noindent
QUANTMOD$\textunderscore$DIRECT$\textunderscore$NUM detects the \emph{change}, by looking at the verb or adjective directly before NUM tokens.
Here, ``fell'' is a trigger word for a downward trend.
For example 1, we thus have two extracted triplets with \emph{value}, \emph{unit}, and \emph{change}.
\begin{itemize}
\item \hspace{-0.2cm} $\langle v=0.4,\ u=pc,\ ch=fell \rangle$
\item\hspace{-0.2cm} $\langle v=0.1,\ u=None,\ ch=gained \rangle$,
\end{itemize}
 In Appendix~\ref{sec:appendix1}, more examples are given.\\
 \noindent
If no \emph{unit} is detected for a quantity, its context is checked for the possibility of \emph{shared units}. For the quantity {\sloppy$\langle v=0.1,\ u=None,\ ch=gained\rangle$} in Example 1 ,``percentage'' is the derived \emph{unit}, although not mentioned in the text.
\emph{Shared units} often occur in similarly structured sub-clauses or after connector words such as ``and'', ``while'', or ``whereas''.
The similarity between two sub-clauses is computed using the \emph{Levenshtein ratio} between the structure of clauses.
The structure is represented by POS-tags, e.g., ``German DAX fell 0.4 pc'' $\to$ ``JJ NNP VBD CD NN'' and ``the CAC40 in France gained 0.1'' $\to$``DT NNP IN NNP VBD CD''.
This ratio is between 0 and 100, where larger values indicate higher similarity.
If connector words are present and ratio is larger than 60, the unitless quantity is assigned the \emph{unit} of the other sub-clause.
e.g., $None$ becomes $pc$.\\
\noindent
Finally, the candidate \emph{values} are filtered by logical rules to avoid false detection of \emph{non-quantities}, e.g., in ``S\&P 500'', 500 is not a quantity.

\subsubsection{Concept Detection}
\emph{Concepts} are detected in one of the following ways, ordered by priority:

\begin{enumerate}
\item  Keywords, such as \emph{for}, \emph{of}, \emph{at} or \emph{by} before or after a \emph{value} point to a potential concept. 
For example, ``with carbon levels \emph{at} 1200 parts per million'' results in $cn=(carbon,\ levels)$.
The noun and pronouns before and after such keywords are  potential concepts.
\item The entire subtree of dependencies with a number as one of the leaf nodes is inspected to find the closest verb related to the number. 
If no verb is found, then the verb connected to the ROOT is selected.
The nominal subject of the verb is considered as the \emph{concept}. 
In Example 1, both ``German DAX'' and ``CAC40 in France'' are the nominal subjects of the closest verbs to the number tokens in the text.
\item Sometimes values occur in a relative clause that modifies the nominal, e.g., ``maximum investment per person, which is 50000'' $\to$ $cn=(maximum,\ investment,\ per,\ person)$. 
In such a case, the noun phrase before the relative clause is the \emph{concept}, since the relative clause is describing it. 

\item If the numerical \emph{value} in a sentence is not associated with the nominal of the sentence, then it is mostly likely related to the object. 
Therefore, the direct object of the verb is also a candidate, e.g., ``She gave me a raise of \$1k'', where ``raise'' is the direct object of the verb.
\item Finally, if the \emph{concept} is not found in the previous step, and there is a single noun in the sentence, the noun is tagged as the \emph{concept}, e.g., ``a beetle that can go from 0 to 80 km/h in about 8 seconds, ''  $\to$ $cn=(beetle)$.
\end{enumerate}
From the list of candidate tokens for \emph{concepts}, tokens previously associated with \emph{units} and \emph{values} are filtered and stopwords  are remove,
e.g.,  ``CAC40 in France'' results$cn=(CAC40,\ France)$.
Generally, a \emph{concept} is represented as a list of tokens.


\subsubsection{Normalization and Standardization}
The final stage is the normalization of \emph{units} and \emph{changes} using dictionaries and standardization of \emph{values}.
The \emph{units dictionary} is a set of 531 \emph{units}, their surface forms and symbols gathered from the Quantulum3 library, a dictionary provided by Unified Code for Units of Measure (UCUM)~\cite{DBLP:conf/esws/LefrancoisZ18}, and a list of \emph{units} from Wikipedia~\footnote{\url{https://en.wikipedia.org/wiki/Template:Convert/list_of_units} Last accessed: April 17, 2023}. 
An example of an entry in this dictionary for ``euro'' is:
\begin{verbatim}
{"euro": 
"surfaces": ["Euro","Euros","euro",
"euros"],
"symbols": ["EUR","eur",€]}}.
\end{verbatim}
\noindent  
The detected token span of a \emph{unit} is normalized by matching against the different surface forms and symbols in the dictionary. 
The normalized form is the key of the dictionary and is added to the output, e.g., ``euro'' in the example above or ``cm'' giving ``centimetre''.
The normalization makes the comparison of different units easier. 
Note that conversions between metric units is not supported. For example, 
``centimetre'' is kept as the final representation and not converted to ``metre''. \\
\noindent
If the detected surface form is shared across multiple \emph{units}, the unit is \emph{ambiguous} and requires further normalization based on the context.
Since language models are great at capturing contextual information, for this purpose, we train a BERT-based classifier~\cite{devlin-etal-2019-bert}. 
There are 18 ambiguous surface forms in our unit dictionary, and for each a separate classifier is trained that allows to distinguish among \emph{units} based on the context. 
If an ambiguous surface form is detected by the system, the relevant classifier is used to find the correct normalized unit. \\ 
\noindent
\emph{Compound units} are also detected and normalized independently. For example, ``kV/cm'' results in ``kilovolt per centimetre', where.
``kV'' and ``cm'' are normalized based on separate dictionary entries.

\noindent
If no valid match in the dictionary exists, the surface form is tagged as a \emph{noun unit} and lemmatizated, e.g., for ``10 students'' gives $u=student$.
In some cases, the adjective before a noun is also part of the unit, e.g., ``two residential suites'' results in $u=residential\ suite$.

\noindent
The \emph{value dictionary} contains the necessary information to standardize  \emph{values} to real numbers. 
More specifically, it contains surface forms for prefixes and suffixes of scales, e.g., ``B: billion'' or ``n: nano'',   
spelled out numbers in textual format, e.g., ``fourty-two: 42'', fractions in textual format, e.g., ``half: 1/2'', and scientific exponents, e.g., ``$10^2$: 100'. This combination is used to convert \emph{values} to decimal format.
Scientific notations with exponent and mantissa are converted to decimal values, e.g.,``$2.3E2$ $\to$ $v=23$''.\\
\noindent
Various trigger words or symbols for bounds and trends are managed in the \emph{changes dictionary}, where detected tokens for \emph{change} are mapped to one of the allowed categories $\sim, =, >, <, up, down$. For example, the entry for equality is 
\texttt{\{"=": [ "exactly", "just", "equals", "totalling","="]}.\\

\vspace*{-0.2cm}
\section{Evaluation}

\begin{table*}[h]
\caption{Comparison of functionality for various extractors. }  
\label{tab:capabilities}
\setlength{\tabcolsep}{20pt} 
\renewcommand{\arraystretch}{1} 
\resizebox{\textwidth}{!}{%
\begin{tabular}{llcccc}
\toprule
 Feature & Example  & CQE&  IllQ & R-Txt & Q3   \\
\midrule
Value  &   5k euros (5k)  & \cmark & \cmark & \cmark   & \cmark  \\
Standardization & 5k euros (5000) & \cmark  & \cmark  & \cmark   & \cmark  \\
Negative Values & -5 C (-5)  &\cmark    & \xmark    & \cmark  &\cmark \\
Fractions & 1/3 of the population (0.33)  &\cmark    & \xmark    & \cmark  &\cmark \\
Range & 40-60 km/h (40-60)  &\cmark    & \xmark    & \xmark  &\cmark \\
Non-quantities & iPhone 11 (-)  &\cmark    & \xmark    & \xmark  &\xmark   \\
Scientific Notation & $1.9\times 10^2$ (190) &\cmark    & \xmark    & \xmark  &\cmark   \\

\midrule
Unit  & 1mm (mm)  & \cmark  & \cmark  & \cmark   & \cmark  \\
Unit normalization & 1mm (millimetre)  & \cmark  & \xmark  & \cmark   & \cmark  \\
Unit disambiguation & 10 pound (sterling or mass?) & \cmark   & \xmark   & \xmark  & \cmark  \\
Noun Units & 200 people (people)  &\cmark    & \cmark   & \xmark  & \xmark  \\
Shared Units & about 8 or \$9 (both dollar)  &\cmark    & \xmark   & \xmark  & \cmark  \\
\midrule
Change  & more than 100 (>)     & \cmark & \cmark & \xmark  & \xmark  \\
Trends & DAX fell 2\% (down)  &\cmark    & \xmark    & \xmark  &\xmark \\
\midrule
Concept  & AAPL rose 2\% (AAPL)  & \cmark  &  \xmark  & \xmark  & \xmark  \\
\bottomrule
\end{tabular}
}
\end{table*}
\begin{table*}[h]
\caption{Statistics of the  number of sentences, quantities, and
  sentences with and without quantities in the NewsQuant and R-Txt datasets.} 
\label{tab:stats}
\setlength{\tabcolsep}{22pt} 
\renewcommand{\arraystretch}{1} 
\resizebox{\textwidth}{!}{%
\begin{tabular}{lccccc}
\toprule
  Dataset & \#sent & \#quantity & \#sent with quantity  & \#sent w/o quantity  \\
\midrule
NewsQuant  & 590 & 904  & 475 & 115\\
R-Txt-currencies   & 180 & 255 & 178 & 2 \\
R-Txt-dimension   & 93 & 121 & 77 & 14 \\
R-Txt-temperature  & 36 & 34 & 34 & 2 \\
R-Txt-age   & 19 & 22 & 18 & 1\\
\bottomrule
\end{tabular}%
}
\end{table*}

CQE is compared against \emph{Illinois Quantifier} (IllQ),
\emph{Quantulum3} (Q3), \emph{Recognizers-Text} (R-Txt), and GPT-3
with few-shot learning~\cite{DBLP:conf/nips/BrownMRSKDNSSAA20}.  From
here on, the abbreviation is used to refer to the respective system.
We first compare the functionality of the models, then describe our
benchmark dataset and compare the models on precision, recall and
F1-score for quantity extraction.  Finally, unit
disambiguation module is evaluated on a custom-made dataset against Q3.  Our
evaluation code and datasets are available at
\url{https://github.com/satya77/CQE_Evaluation}.

\subsection{Comparison of Functionality}

Table~\ref{tab:capabilities} compares the functionality of the models
in terms of different types of \emph{values}, \emph{units} and 
\emph{changes}, as well as normalization techniques.  IllQ is the only
baseline to detect \emph{changes} in \emph{values} but in a limited
setting that does not consider upward or downward trends.  IllQ
performs normalization for currencies, however, \emph{scientific
  units} are not normalized.  Furthermore, it fails to detect
\emph{fractional values}.  After our approach (CQE), Q3 has the most
functionality and is the only model that correctly detects
\emph{ranges} and \emph{shared units} and performs \emph{unit}
disambiguation.  On the other hand, Q3 disregards \emph{noun-based
  units}, and although it is capable of detecting a wide range of
\emph{value} types, it makes incorrect detections of
\emph{non-quantitative values}.  R-Txt has dedicated models for
certain quantity types but fails to detect other types in the text.
GPT-3 has a lot of variability in the output and does not provide
concrete and stable functionality like the models discussed in this
section, therefore, it is not further considered in this comparison.

\subsection{NewsQuant Dataset}

For a quantitative comparison, we introduce a new evaluation resource
called NewsQuant, consisting of 590 sentences from news articles in
the domains of economics, sports, technology, cars, science, and
companies.  To the best of our knowledge, this is the first
comprehensive evaluation set introduced for quantity extraction.  Each
sentence is tagged with one or more quantities containing
\emph{value}, \emph{unit}, \emph{change}, and \emph{concept} and is
annotated by the two first authors of the paper. 
Inter-annotator agreements separately for \emph{value}, \emph{unit},
\emph{change}, and \emph{concept} is computed.  For the first three, the Cohen Kappa
coefficient~\cite{cohen1960coefficient} with values of $1.0$, $0.92$,
and $0.85$ is reported.  Value detection is an easy task for humans
and annotators have perfect agreement.  A \emph{concept} is a span of
tokens in the text and does not have a standardized representation,
therefore, Cohen Kappa coefficient can not be used and instead
Krippendorff’s alpha~\cite{krippendorff2011computing}, with the value
of $0.79$, is reported.  A set of guidelines was designed for what constitutes a
concept of higher quality. In total, the annotators completely agreed
on all elements for $62\%$ of the annotations. \\
\noindent
We additionally evaluate four datasets available in the repository of
R-Txt for age, dimension, temperature, and
currencies\footnote{\url{https://github.com/microsoft/Recognizers-Text/tree/master/Specs/NumberWithUnit/English}
  Last accessed: May 4, 2023}.  These datasets  contain only
\emph{unit/value} pairs. The original datasets only
contained tags for the certain quantity type and would ignore other
types, giving the R-Txt model an advantage. For example, in the
R-Txt-currencies, only the currencies were annotated and other
quantities were ignored.  We added extra annotations for all other
types of quantities for a fair comparison.  For example, in the sentence ``I
want to earn \$10000 in 3 years'', where only ``\$10000'' was
annotated, we add ``3 years''.  Statistics of the number of sentences
and quantities for each dataset are shown in Table ~\ref{tab:stats}.
The NewsQuant dataset is the largest dataset for this task containing
over 900 quantities of various types.  NewsQuant also includes
negative examples with non-quantity numerals.
 
\subsection{Disambiguation Dataset}

To train our unit disambiguation system, a dataset of 18
ambiguous surface forms is created, using
ChatGPT\footnote{\url{https://chat.openai.com/} \sloppy{Last accessed: May 4, 2023}}. For
each ambiguous surface form, at least 100 examples are generated, and
the final training dataset consists of 1,835 sentences with various
context information.  For more challenging surface forms, more samples
are generated.  For the list of ambiguous surface forms and the number
of samples for each class, refer to Appendix~\ref{sec:appendix3}.  A
test dataset is generated in the same manner using ChatGPT, consisting
of 180 samples, 10 samples per surface form.  For more information on the
dataset creation, please see
Appendix~\ref{sec:appendix4}.

\begin{table*}[t]
\caption{Precision, recall, and F1-score for detection of \emph{value}, \emph{unit} and \emph{change} on NewsQuant.  } 
\label{tab:valueUnitChange}
\setlength{\tabcolsep}{16pt} 
\renewcommand{\arraystretch}{1} 
\resizebox{\textwidth}{!}{%
\begin{tabular}{lccc|ccc|ccc}
\toprule
  \multicolumn{1}{c}{\multirow{2}{*}{Model}}&\multicolumn{3}{c}{Value}& \multicolumn{3}{c}{Value+Unit}&\multicolumn{3}{c}{Value+Change}  \\
\cmidrule{2-10}
 \multicolumn{1}{c}{}  & P&R&\multicolumn{1}{c}{F1} & P&R&\multicolumn{1}{c}{F1}  & P&R &F1  \\

\midrule
CQE   &$\mathbf{92.0}$ &	$\mathbf{91.9}$&	$\mathbf{92.0}^{\dagger}$   & $\mathbf{85.6}$	&$\mathbf{85.5}$	&$\mathbf{85.6}^{\dagger}$ &  $\mathbf{88.2}$	&$\mathbf{88.1}$	&$\mathbf{88.1}^{\dagger}$ \\
Q3   & 65.0	&83.3&	73.0  &  42.1&	53.9	&47.2& - & - & -   \\
IllQ  & 50.6&	66.0&	57.3 &   32.8&	42.8&	37.1  &  44.2&	57.6&	50.0    \\
R-Txt   & 59.7 &	82.2&	69.1    &29.6	&40.7&	34.2& - & - & -  \\
GPT-3   & 72.1	&69.1&	70.6  &  60.3&	57.9&	59.1&    53.1&	50.9	&51.9 \\
\bottomrule
\end{tabular}%
}
\end{table*}

\begin{table*}[t]
\caption{Precision, recall and F1-score for detection of \emph{value} and \emph{unit} on R-Txt Datasets.  } 
\label{tab:valueUnit}
\setlength{\tabcolsep}{8pt} 
\renewcommand{\arraystretch}{1} 
\resizebox{\textwidth}{!}{%
\begin{tabular}{l|l|ccc|ccc|ccc|ccc}
\toprule
\multicolumn{1}{c}{\multirow{2}{*}{Model}} &\multicolumn{1}{c}{\multirow{2}{*}{Detect}} & \multicolumn{3}{c}{currency} & \multicolumn{3}{c}{dimension} & \multicolumn{3}{c}{temperature}& \multicolumn{3}{c}{age}  \\
\cmidrule{3-14}
  \multicolumn{1}{c}{} &\multicolumn{1}{c}{} & P&R&\multicolumn{1}{c}{F1} & P&R&\multicolumn{1}{c}{F1}  & P&R &\multicolumn{1}{c}{F1}  & P&R &F1 \\
\midrule

CQE & \multirow{5}{*}{Value} & $\mathbf{82.6}$	 & 85.9	& $\mathbf{84.2}$ &$\mathbf{85.5}$	& 87.6&	$\mathbf{86.5}$ &$\mathbf{94.3}$	&97.1&	95.7&$\mathbf{91.3}$	&$\mathbf{95.5}$	&$\mathbf{93.3}$  \\
Q3  & & 69.2 &84.7 &76.2 &76.9 &$\mathbf{93.4}$ &84.3 &91.7 &97.1 &94.3 &$\mathbf{91.3}$ &$\mathbf{95.5}$ &$\mathbf{93.3}$  \\
IllQ & & 65.5 &70.6 &67.9 &65.3 &77.7 &70.9 &88.9 &94.1 &91.4 &65.4 &77.3 &70.8 \\
R-Txt  & & 67.4 &$\mathbf{91.8}$ &77.7 &73.6 &90.1 &81.0 &91.9 &$\mathbf{100.0}$ &\textbf{95.8} &77.8 &$\mathbf{95.5}$ &85.7   \\
GPT-3  & & 50.5 &54.9 &52.6 &80.2 &80.2 &80.2 &93.5 &85.3 &89.2 &92.3 &54.5 &68.6  \\
\midrule
\midrule
CQE &  & $\mathbf{78.1}$	&$\mathbf{81.2}$	& $\mathbf{79.6}^{\dagger}$ &$\mathbf{78.2}$	&$\mathbf{80.2}$&	$\mathbf{79.2}$ &91.4	&94.1&	92.8 &$\mathbf{91.3}$	&$\mathbf{95.5}$	&$\mathbf{93.3}$\\
Q3  & Value & 29.5 &36.1 &32.5 &56.5&	68.6&	61.9 &61.1 &76.5 &74.3&82.6 &86.4 &84.4  \\
IllQ & +Unit & 41.8 &41.6 &45.1 &43.4 &52.1 &47.5 &30.6 &32.4 &31.4 &42.3 &50.0 &45.8 \\
R-Txt & &46.7 &63.5 &53.8 &44.6 &54.5 &49.1&$\mathbf{91.9}$ &$\mathbf{100.0}$ &$\mathbf{95.8}$ &70.4 &86.4 &77.6   \\
GPT-3  & & 40.8 &44.3 &42.5 &65.3 &65.3 &65.3 &45.2 &41.2 &43.1 &92.3 &54.5 &68.6 \\
\bottomrule
\end{tabular}%
}
\end{table*}

\begin{table}[t]
\caption{Relaxed and strict matching, precision, recall and F1-score for \emph{concept} detection on the NewsQuant.} 
\label{tab:concept}
\setlength{\tabcolsep}{8pt} 
\renewcommand{\arraystretch}{1} 
\resizebox{0.5\textwidth}{!}{%
\begin{tabular}{lccc|ccc}
\toprule
 \multicolumn{1}{c}{\multirow{2}{*}{Model}} &\multicolumn{3}{c}{ Relaxed Match}& \multicolumn{3}{c}{Strict Match}  \\
 \cmidrule{2-7}
\multicolumn{1}{c}{}   & P&R&\multicolumn{1}{c}{F1} & P&R&F1   \\

\midrule
		    		
CQE   & $\mathbf{76.2}$ &$\mathbf{76.1}$ &$\mathbf{76.1^{\dagger}}$ &$\mathbf{57.0}$ &$\mathbf{57.0}$ &$\mathbf{57.0^{\dagger}}$   \\
GPT-3   & 55.9	&53.7	&54.8   & 26.3&	25.2&	25.7    \\
\bottomrule
\end{tabular}%
}
\end{table}

\begin{table}[t]
\caption{Weighted micro-average precision, recall and F1-score on the \emph{unit} disambiguation dataset.} 
\label{tab:disam}
\setlength{\tabcolsep}{23pt} 
\renewcommand{\arraystretch}{1} 
\resizebox{0.5\textwidth}{!}{%
\begin{tabular}{lccc}
\toprule
Model  & P&R&F1 \\
\midrule
CQE   & $\mathbf{89.9}$ &$\mathbf{89.4}$ &$\mathbf{88.1^{\dagger}}$   \\
Q3   & 57.33 &57.78 &54.46    \\
\bottomrule
\end{tabular}%
}
\end{table}

\subsection{Implementation}
CQE is implemented in Python 3.10.  For dependency parsing,
part-of-speech tagging, and the matching of rules SpaCy
3.0.9\footnote{\url{https://spacy.io/} Last accessed: May 4, 2023} is used.  The unit
disambiguation module, with BERT-based classifiers, is trained using
spacy-transformers\footnote{\url{https://spacy.io/universe/project/spacy-transformers}
  Last accessed: May 4, 2023} for smooth intergeneration with other SpaCy
modules.
%
Parsers were created to align the output format of different baselines
so that the differences in output representation do not affect the
evaluation.  For instance, for IllQ, we normalize the \emph{scientific
  units} and account for differences in the representation of ranges
in Q3. If a value is detected by a baseline but not standardized or a
unit is not normalized to the form present in the dataset,
post-processing is applied for a unified output. \\
%
Moreover, to keep up with the recent trends in NLP and the lack of a
baseline for \emph{concept} detection, we introduce a GPT-3 baseline.
The GPT-3 model is prompted to tag quantities with 10 examples for
few-shot learning. Prompts and examples are available in our
repository.  We use the \textit{ text-davinci-003} model from the
GPT-3 API\footnote{\url{https://platform.openai.com/} Last accessed: February 2,
  2023} with a sequence length of 512, temperature of 0.5, and no
frequency or presence penalty.  For more details, refer to
Appendix~\ref{sec:appendix2}. We are aware that with extensive
fine-tuning and more training examples GPT-3 values are likely to
improve.  However, the purpose of this paper is neither prompt
engineering nor designing training data for GPT-3, and the few-short
learning should suffice for a baseline.

\subsection{Analysis of Results}
\noindent
All the models are compared on precision, recall, and F1-score for the
detection of \emph{value}, \emph{unit}, \emph{change}, and
\emph{concept}.  Disambiguation systems are also compared regarding
precision, recall, and F1-score of unit classification.
Permutation re-sampling is used to test for significant improvements in
F1-scores~\cite{riezler-maxwell-2005-pitfalls}, which is more
statistically coherent in comparison to the commonly paired bootstrap
sampling~\cite{DBLP:conf/emnlp/Koehn04}.
Results denoted with $\dagger$ mark highly significant improvements
over the best-performing baseline with a $p$-value < 0.01.

\noindent
\textbf{NewsQuant:} Table ~\ref{tab:valueUnitChange} shows the result
on the NewsQuant dataset. Since Q3 and R-Txt do not detect changes,
respective entries are left empty. CQE beats all baselines in each
category by a significant margin, where most of the errors are due to
incorrect extraction of the dependency parsing tree and part-of-speech
tagging.  The second best model, Q3, scores highly for \emph{value}
detection, but misses all the noun base \emph{units} and tends to
overgeneralize tokens to \emph{units} where none exist, e.g., in ``0.1
percent at 5884'', Q3, detects \emph{percent per ampere-turn} as a
\emph{unit}.  Moreover, Q3 makes mistakes on different currencies and
their normalization. We attribute this to their incomplete unit
dictionary.  R-Txt works well for the quantity types with dedicated
models, but all the other quantities are ignored or misclassified.
IllQ has trouble with \emph{compound units}, e.g., ``\$2.1 per
gallon'' and tends to tag the word after a \emph{value} as unit, e.g.,
in ``women aged 25 to 54 grew by 1\%'', \emph{grew by} is the detected
\emph{unit}.  Although IllQ is supposed to normalize currencies, in
practice the normalization is limited and often currency symbols are
not normalized.  Moreover, trends are ignored by IllQ, and the model
is biased to predict equality (\emph{=}) for most \emph{changes} and
other change types are rare.  GPT-3 achieves a score close to Q3 for
the detection of \emph{units} and \emph{values} and close to IllQ for
\emph{changes}.  Nevertheless, due to extreme hallucination, extensive
post-processing of the output is required for evaluation, e.g., many
of the values extracted were not actual numbers and units are not
normalized.  Moreover, GPT-3 often confuses value suffixes with
\emph{units}, e.g., ``billion'' or ``million'' and, despite the
normalization prompt, fails to normalize \emph{units} and required
manual normalization for most detections.  Both IllQ and GPT-3 require
extensive post-processing for units and cannot easily be used outside
of the box.

\noindent 
\textbf{R-Txt Dataset:} Evaluation results on the four quantity types
of the R-Txt dataset are shown in Table~\ref{tab:valueUnit}, where our
model once again outperforms all baselines on \emph{value+unit}
detection for all categories except for temperature.  Nevertheless,
for temperature, the R-Txt improvement over CQE is not statistically
significant. 
Small size of the age and temperature dataset results in inconsistent significance testing.
The closeness of \emph{value}
detection between models is due to the structure of the
dataset. \emph{Value}s are floats, and the diversity of types
like ranges, fractions, and non-quantities is negligible. For more
details in the error analysis on NewsQuant and R-Txt, see
Appendix~\ref{sec:appendix6}.


\noindent 
\textbf{Concept Detection:} Finally, \emph{concept}
detection is evaluated on NewsQuant dataset. Results are shown in
Table~\ref{tab:concept}. Following the approach of UzZaman et
al.~\cite{uzzaman2013semeval} for evaluation, strict
and relaxed matches are compared. A strict match is an exact token match,
whereas a relaxed match is counted when there is an overlap between
the system's and ground truth token spans.  Based on the scores we
observe that \emph{concept} detection is harder in comparison to
\emph{value+unit} detection.  Even GPT-3 struggles with accurate
predictions. Our algorithm for \emph{concept} detection is limited to
common cases and does not take into account the full complexity of
human language, leaving room for improvement in future work.
Moreover, in many cases, the concept is implicit and hard to
distinguish even for human annotators.  In general, our approach is
more recall oriented, trying to capture as many concepts as possible,
hence, the big gap between partial and complete matches.  However,
since the method is rule-based, rules can be adjusted to be
restrictive and precision focused.

\noindent 
\textbf{Unit Disambiguation:} CQE is compared against Q3 (the only other
systems with disambiguation capabilities) in Table~\ref{tab:disam}.
Since the normalization of a \emph{units} is not consistent in the
GPT-3 model and requires manual normalization, GPT-3 is left out of
this study.  All 18 classifiers are evaluated within a single system.
The results are averaged by weighting the score of each class label by
the number of true instances when calculating the average.  CQE
significantly outperforms Q3 on all metrics, and it is easily
expendable to new surface forms and \emph{units} by adding a new
classifier.  Since the training data is generated using ChatGPT, a new
classifier can be trained using our paradigm and data generation
steps, as shown in in Appendix~\ref{sec:appendix4}.  For a detailed
evaluation of each class, see Appendix~\ref{sec:appendix5}.

\label{sec:eval}

\section{Conclusion and Ongoing Work}
\label{sec:conclusion}
In this paper, we introduced CQE, a comprehensive quantity extractor for unstructured text.  Our system is not only significantly beating related methods and frameworks as well as a GPT-3 neural model for the detection of \emph{values}, \emph{units} and \emph{changes} but also introduces the novel task of \emph{concept} detection.  Furthermore, we present the first benchmark dataset for the comprehensive evaluation of quantity extraction and make our code and data available to the community. We are currently extending the extractor by improving the quality of edge cases.

\bibliography{anthology,custom}

\begin{thebibliography}{20}
\expandafter\ifx\csname natexlab\endcsname\relax\def\natexlab#1{#1}\fi

\bibitem[{Almasian et~al.(2022)Almasian, Bruseva, and
  Gertz}]{DBLP:conf/sigir/AlmasianB022}
Satya Almasian, Milena Bruseva, and Michael Gertz. 2022.
\newblock Qfinder: {A} framework for quantity-centric ranking.
\newblock In \emph{{SIGIR} '22: The 45th International {ACM} {SIGIR} Conference
  on Research and Development in Information Retrieval, Madrid, Spain, July 11
  - 15, 2022}, pages 3272--3277. {ACM}.

\bibitem[{Banerjee et~al.(2009)Banerjee, Chakrabarti, and
  Ramakrishnan}]{DBLP:conf/sigir/BanerjeeCR09}
Somnath Banerjee, Soumen Chakrabarti, and Ganesh Ramakrishnan. 2009.
\newblock {Learning to Rank for Quantity Consensus Queries}.
\newblock In \emph{Proceedings of the 32nd Annual International {ACM} {SIGIR}
  Conference on Research and Development in Information Retrieval, {SIGIR}
  2009, Boston, MA, USA, July 19-23, 2009}, pages 243--250. {ACM}.

\bibitem[{Brown et~al.(2020)Brown, Mann, Ryder, Subbiah, Kaplan, Dhariwal,
  Neelakantan, Shyam, Sastry, Askell, Agarwal, Herbert{-}Voss, Krueger,
  Henighan, Child, Ramesh, Ziegler, Wu, Winter, Hesse, Chen, Sigler, Litwin,
  Gray, Chess, Clark, Berner, McCandlish, Radford, Sutskever, and
  Amodei}]{DBLP:conf/nips/BrownMRSKDNSSAA20}
Tom~B. Brown, Benjamin Mann, Nick Ryder, Melanie Subbiah, Jared Kaplan,
  Prafulla Dhariwal, Arvind Neelakantan, Pranav Shyam, Girish Sastry, Amanda
  Askell, Sandhini Agarwal, Ariel Herbert{-}Voss, Gretchen Krueger, Tom
  Henighan, Rewon Child, Aditya Ramesh, Daniel~M. Ziegler, Jeffrey Wu, Clemens
  Winter, Christopher Hesse, Mark Chen, Eric Sigler, Mateusz Litwin, Scott
  Gray, Benjamin Chess, Jack Clark, Christopher Berner, Sam McCandlish, Alec
  Radford, Ilya Sutskever, and Dario Amodei. 2020.
\newblock Language models are few-shot learners.
\newblock In \emph{Advances in Neural Information Processing Systems 33: Annual
  Conference on Neural Information Processing Systems 2020, NeurIPS 2020,
  December 6-12, 2020, virtual}.

\bibitem[{Chen et~al.(2023)Chen, Chen, and Karlsson}]{chen2023dataset}
Sanxing Chen, Yongqiang Chen, and B{\"o}rje~F Karlsson. 2023.
\newblock Dataset and baseline system for multi-lingual extraction and
  normalization of temporal and numerical expressions.
\newblock \emph{arXiv preprint arXiv:2303.18103}.

\bibitem[{Cohen(1960)}]{cohen1960coefficient}
Jacob Cohen. 1960.
\newblock A coefficient of agreement for nominal scales.
\newblock \emph{Educational and Psychological Measurement}, 20(1):37--46.

\bibitem[{Devlin et~al.(2019)Devlin, Chang, Lee, and
  Toutanova}]{devlin-etal-2019-bert}
Jacob Devlin, Ming-Wei Chang, Kenton Lee, and Kristina Toutanova. 2019.
\newblock \href {https://doi.org/10.18653/v1/N19-1423} {{BERT}: Pre-training of
  deep bidirectional transformers for language understanding}.
\newblock In \emph{Proceedings of the 2019 Conference of the North {A}merican
  Chapter of the Association for Computational Linguistics: Human Language
  Technologies, Volume 1 (Long and Short Papers)}, pages 4171--4186,
  Minneapolis, Minnesota. Association for Computational Linguistics.

\bibitem[{Foppiano et~al.(2019)Foppiano, Romary, Ishii, and
  Tanifuji}]{DBLP:conf/doceng/FoppianoRIT19}
Luca Foppiano, Laurent Romary, Masashi Ishii, and Mikiko Tanifuji. 2019.
\newblock Automatic identification and normalisation of physical measurements
  in scientific literature.
\newblock In \emph{Proceedings of the {ACM} Symposium on Document Engineering
  2019, Berlin, Germany, September 23-26, 2019}, pages 24:1--24:4. {ACM}.

\bibitem[{Forbus(1984)}]{DBLP:journals/ai/Forbus84}
Kenneth~D. Forbus. 1984.
\newblock Qualitative process theory.
\newblock \emph{Artif. Intell.}, 24(1-3):85--168.

\bibitem[{Ho et~al.(2019)Ho, Ibrahim, Pal, Berberich, and
  Weikum}]{DBLP:conf/semweb/HoIPBW19}
Vinh~Thinh Ho, Yusra Ibrahim, Koninika Pal, Klaus Berberich, and Gerhard
  Weikum. 2019.
\newblock Qsearch: Answering quantity queries from text.
\newblock In \emph{The Semantic Web - {ISWC} 2019 - 18th International Semantic
  Web Conference, Auckland, New Zealand, October 26-30, 2019, Proceedings, Part
  {I}}, volume 11778 of \emph{Lecture Notes in Computer Science}, pages
  237--257. Springer.

\bibitem[{Huang et~al.(2017)Huang, Lin, McConnell, and
  Karlsson}]{soft:recognizers-text}
Wenhao Huang, Zijia Lin, Chris McConnell, and B{\"{o}}rje~F. Karlsson. 2017.
\newblock {Recognizers-Text: {R}ecognition and Resolution of Numbers, Units,
  and Date/time Entities Expressed Across Multiple Languages}.

\bibitem[{Koehn(2004)}]{DBLP:conf/emnlp/Koehn04}
Philipp Koehn. 2004.
\newblock Statistical significance tests for machine translation evaluation.
\newblock In \emph{Proceedings of the 2004 Conference on Empirical Methods in
  Natural Language Processing , {EMNLP} 2004, {A} meeting of SIGDAT, a Special
  Interest Group of the ACL, held in conjunction with {ACL} 2004, 25-26 July
  2004, Barcelona, Spain}, pages 388--395. {ACL}.

\bibitem[{Krippendorff(2011)}]{krippendorff2011computing}
Klaus Krippendorff. 2011.
\newblock Computing krippendorff's alpha-reliability.

\bibitem[{Lefran{\c{c}}ois and Zimmermann(2018)}]{DBLP:conf/esws/LefrancoisZ18}
Maxime Lefran{\c{c}}ois and Antoine Zimmermann. 2018.
\newblock The unified code for units of measure in {RDF:} cdt: ucum and other
  {UCUM} datatypes.
\newblock In \emph{The Semantic Web: {ESWC} 2018 Satellite Events - {ESWC} 2018
  Satellite Events, Heraklion, Crete, Greece, June 3-7, 2018, Revised Selected
  Papers}, volume 11155 of \emph{Lecture Notes in Computer Science}, pages
  196--201. Springer.

\bibitem[{Li et~al.(2021)Li, Fang, Lou, Li, and
  Zhang}]{DBLP:conf/wsdm/LiFLLZ21}
Tongliang Li, Lei Fang, Jian{-}Guang Lou, Zhoujun Li, and Dongmei Zhang. 2021.
\newblock {AnaSearch: Extract, Retrieve and Visualize Structured Results from
  Unstructured Text for Analytical Queries}.
\newblock In \emph{{WSDM} '21, The Fourteenth {ACM} International Conference on
  Web Search and Data Mining, Virtual Event, Israel, March 8-12, 2021}, pages
  906--909. {ACM}.

\bibitem[{Maiya et~al.(2015)Maiya, Visser, and Wan}]{DBLP:conf/sigir/MaiyaVW15}
Arun~S. Maiya, Dale Visser, and Andrew Wan. 2015.
\newblock \href {https://doi.org/10.1145/2766462.2767789} {{Mining Measured
  Information from Text}}.
\newblock In \emph{Proceedings of the 38th International {SIGIR} Conference on
  Research and Development in Information Retrieval}, pages 899--902. {ACM}.

\bibitem[{Pennington et~al.(2014)Pennington, Socher, and
  Manning}]{pennington-etal-2014-glove}
Jeffrey Pennington, Richard Socher, and Christopher Manning. 2014.
\newblock \href {https://doi.org/10.3115/v1/D14-1162} {{G}lo{V}e: Global
  vectors for word representation}.
\newblock In \emph{Proceedings of the 2014 Conference on Empirical Methods in
  Natural Language Processing ({EMNLP})}, pages 1532--1543, Doha, Qatar.
  Association for Computational Linguistics.

\bibitem[{Riezler and Maxwell(2005)}]{riezler-maxwell-2005-pitfalls}
Stefan Riezler and John~T. Maxwell. 2005.
\newblock \href {https://aclanthology.org/W05-0908} {On some pitfalls in
  automatic evaluation and significance testing for {MT}}.
\newblock In \emph{Proceedings of the {ACL} Workshop on Intrinsic and Extrinsic
  Evaluation Measures for Machine Translation and/or Summarization}, pages
  57--64, Ann Arbor, Michigan. Association for Computational Linguistics.

\bibitem[{Roy et~al.(2015)Roy, Vieira, and Roth}]{roy-etal-2015-reasoning}
Subhro Roy, Tim Vieira, and Dan Roth. 2015.
\newblock \href {https://doi.org/10.1162/tacl_a_00118} {Reasoning about
  quantities in natural language}.
\newblock \emph{Transactions of the Association for Computational Linguistics},
  3:1--13.

\bibitem[{Sarawagi and Chakrabarti(2014)}]{DBLP:conf/kdd/SarawagiC14}
Sunita Sarawagi and Soumen Chakrabarti. 2014.
\newblock {Open-domain Quantity Queries on Web Tables: Annotation, Response,
  and Consensus Models}.
\newblock In \emph{The 20th {ACM} {SIGKDD} International Conference on
  Knowledge Discovery and Data Mining, {KDD} '14, New York, NY, {USA} - August
  24 - 27, 2014}, pages 711--720. {ACM}.

\bibitem[{UzZaman et~al.(2013)UzZaman, Llorens, Derczynski, Allen, Verhagen,
  and Pustejovsky}]{uzzaman2013semeval}
Naushad UzZaman, Hector Llorens, Leon Derczynski, James Allen, Marc Verhagen,
  and James Pustejovsky. 2013.
\newblock {Semeval-2013 task 1: Tempeval-3: Evaluating Time Expressions,
  Events, and Temporal Relations}.
\newblock In \emph{Second Joint Conference on Lexical and Computational
  Semantics (* SEM), Volume 2: Proceedings of the Seventh International
  Workshop on Semantic Evaluation (SemEval 2013)}, pages 1--9.

\end{thebibliography}
\bibliographystyle{acl_natbib}
\clearpage
\appendix

\section{Appendix}
\label{sec:appendix}
\subsection{Value, Unit and Change Detection Rule}\label{sec:appendix1}
In this section, we provide two additional examples for \emph{value},
\emph{unit}, and \emph{change} detection and describe the logic behind
a few other rules.  

\noindent
Example 2: ``The Meged field has produced in the past about 1 million
barrels of oil, but its last well was capped due to technical problems
that have not been resolved.''

\begin{itemize}
\item NUM$\textunderscore$NUM detects the compound number of
  $1\ million$, where $1$, a number, is the child of $million$, a
  noun, in the dependency tree.
\item QUANTMOD$\textunderscore$DIRECT$\textunderscore$NUM detects the
  relation between the adjective ``about'' to the value $1$, which is
  later identified as the change.
\item
  NOUN$\textunderscore$NUM$\textunderscore$ADP$\textunderscore$RIGHT$\textunderscore$NOUN
  finds a noun or proper noun that has a number as a child in the
  dependency tree.  If there are prepositions in children of the noun,
  they are also considered part of the unit.  In this case,
  $[million,\ barrels,\ of,\ oil]$ are detected using this rule.
\end{itemize}
\noindent 
The naming of the rules is preserved in the repository.  From the
combination of all rules, the candidate tokens 
$[about,\ 1,\ million,\ barrels,\ of,\ oil]$ are extracted as
\emph{value}, \emph{unit}, and \emph{change}.

\noindent
Example 3: ``They have a \$3500 a month mortgage and two kids in private school.''

\begin{itemize}
\item NUM$\textunderscore$SYMBOL matches a symbol followed by a
  number.  In this case, $\$3500$ is detected.
\item NOUN$\textunderscore$NUM$\textunderscore$QUANT finds a number
  with a noun or an adverb as its head in the dependency tree. Here,
  we have $[mortgage,\ 3500,\ \$,\ month]$.
\item UNIT$\textunderscore$FRAC$\textunderscore$2 finds compound units
  with ``per'', ``a'' or ``an'' in between, e.g., $[3500,\ month,\ a]$
\item NOUN$\textunderscore$NUM detects a noun that has a number as a
  child, e.g., $[kids,\ two]$.
\end{itemize}
The mentioned rules contribute to extraction of two candidate quantities:
$[[\$,\ 3500,\ a, month,\ mortgage], [two,\ kids]]$.

\subsection{GPT-3 and Few-shot Learning}\label{sec:appendix2}

To tag sentences using GPT-3, we use the few-shot learning paradigm by
prompting the model to tag quantities and units in the text, given 10
distinct examples.  GPT-3 is mainly advertised as a task-agnostic,
few-shot learner, and we have not performed extensive fine-tuning.
With the 10 examples, we aim to account for a variety of outputs,
e.g., compound units, when no quantity is present, noun-based units,
and prefixes for scaling the magnitude of a value.  Our full prompt is
as follows, where the quantities are outputted in a numbered list,
with an order of \emph{change}, \emph{value}, unit surface form,
\emph{unit}, \emph{concept}.
The unit surface form is used in post-processing in case GPT-3 is not able
to normalize the unit.  

\begin{verbatim}
Tag quantities and units in the texts: 

Sentence: Woot is selling refurbished, 
unlocked iPhone XR phones with 64GB of 
storage for about $330. 
Answer: 
1. =, 1.64, GB, gigabyte, storage  
2. ~, 330, $, dollar, iPhone XR phones 

Sentence: The chain operates more than 600 
supermarkets and less than 800 convenience 
stores.  
Answer: 
1. >, 600, supermarkets, supermarkets, 
chain  
2. <, 800, convenience stores, convenience 
stores, chain  

Sentence: The spacecraft, which is about 
the size of a school bus, flew into Dimorphos 
at a speed of about 4.1 miles per second, 
that's roughly 14,760 
miles per hour (23,760 kilometers per hour).
Answer: 
1. ~, 1.4.1,  miles per second, mile per 
second, spacecraft 
2. ~, 14760, miles per hour, mile per hour, 
spacecraft 
3. ~, 23760, kilometers per hour, kilometer 
per hour, spacecraft  

Sentence: And overnight dogecoin fell from 
0.317 to 0.308, a 2.8 percent drop.   
Answer: 
1. =, 1.0.317-0.308, -, -, dogecoin 
2. =, 2.8, percent, percentage, dogecoin  
Sentence: This is about minus 387 
Fahrenheit (minus 233 Celsius).   
Answer: 
1. ~, -387, Fahrenheit, Fahrenheit, - 
2. ~, -233, Celsius, Celsius, -  

Sentence: WhatsApp more than 2 billion 
users send  fewer than 100bn messages a day.  
Answer: 
1. >, 2000000000, users, users, WhatsApp 
2. <, 100000000000, messages, messages, 
users  

Sentence: This includes colors between red 
and blue - wavelengths ranging between 390 
and 700 nm.   
Answer: 
1. =, 390-700, nm, nanometer, wavelengths  
 
Sentence: You don't have a two-year 
bachelor's degree or a six to eight-year
 phd degree.   
Answer: 
1. =, 2, year, year, bachelors degree 
2. =, 6-8, year, year, phd degree  

Sentence: The price CO2 and fuel 
consumption are not clear. 
Answer: 
No quantities or units

Sentence:{sentence}
Answer:
   \end{verbatim}                 %
\{sentence\} is replaced with the query sentence to be tagged.                      
Nevertheless, the output of GPT-3 is not consistent and requires extreme post-processing. 
The post-processing includes cleaning the predicted values to only
include numbers, normalization of the units even if the unit is
miss-spelled, e.g., ``celsiu'' instead of ``celsius'', ``ppb'' to
``parts-per-billion'', or ``\texteuro '' to ``euro''. 
  \begin{table}[t]
\caption{Ambiguous surface forms, \emph{units} associate with them and the number of samples in the training set for each surface form and unit pair. } 
\label{tab:surfaceforms}
\setlength{\tabcolsep}{8pt} 
\renewcommand{\arraystretch}{1} 
\resizebox{0.5\textwidth}{!}{%
\begin{tabular}{llc}
\toprule
Surface   & Units &\# samples \\
\midrule
 c& cent, celsius & 144  \\
\textyen & chinese yuan, japanese yen &100  \\
kn& croatian kuna, knot &116  \\
p& point, penny& 149  \\
R& south african rand, roentgen &100  \\
b& barn, bit &127  \\
'& foot, minute& 104 \\
$'$& foot, minute& 104 \\
"& inch, second &112  \\
 $"$ & inch, second &112 \\
C& celsius, coulomb& 116  \\
F& fahrenheit, farad &100  \\
kt& kiloton, knot &100 \\
B& byte, bel& 107 \\
P& poise, pixel& 102 \\
dram& armenian dram, dram &180 \\
pound& pound sterling, pound-mass& 131 \\
a& acre, year& 113 \\
\bottomrule
\end{tabular}%
}
\end{table}
\begin{table*}[t]
\caption{Error analysis of different extraction systems.} 
\label{tab:mistakes}
\setlength{\tabcolsep}{12pt} 
\renewcommand{\arraystretch}{1} 
\resizebox{\textwidth}{!}{%
\begin{tabular}{ll}
\toprule
Mistake   & Systems  \\
\midrule
 Trouble detecting temperature types, e.g., celsius and fahrenheit are both denoted as degree. & Q3, IllQ, GPT-3\\
 Dollar types are not identified, e.g.,``hong kong dollar'' and ``new zealand dollar'' $\to$ dollar. & Q3, IllQ\\
 Unit normalization does not work for the majority of the times. & IllQ, GPT-3\\
 Bias towards predicting $=$ for changes. & IllQ, GPT-3\\
 Cryptocurrencies and rare ones are not recognized, e.g. Bitcoin or Markka. & Q3 \\
 Sports units are not recognized, e.g., ppg, rpg, apg. & Q3 \\
 Temporal values are mistaken as quantities, e.g., 2 pm. & Txt-R, GPT-3\\
 Compound units are rarely found, e.g., kph. & Txt-R, GPT-3\\
 Units in short sentences are not recognized, e.g., rmb 10 usd 20. & CQE \\
 Problematic distinction between ``year'' and ``year of age''. & CQE, Q3, GPT-3\\
Units are confused with concepts, e.g., building rate of \$80 per sq m $\to$ ``sq m'' as a concept. & GPT-3 \\
Low recall due to limited quantity types. & Txt-R \\
Detection of concepts where none exist. & CQE\\
\bottomrule
\end{tabular}%
}
\end{table*}
\subsection{Ambiguous Surface Forms}\label{sec:appendix3}

In our unit dictionary, we encountered 18 ambiguous surface forms with
different normalized units and collected at least 100 samples for
each. This list is not comprehensive and in different scientific
domains, more ambiguous cases might occur.  The number of samples per
surface form and associated units for each surface form are shown in
Table~\ref{tab:surfaceforms}.

\subsection{Disambiguation Prompts}\label{sec:appendix4}

To generate the dataset for disambiguation, we experimented with
multiple prompts, using ChatGPT.  The aim was to create training/test
data in JSON-format, where the sentences are not duplicates or too
simple.  For this purpose two sentences were formulated (one for each
unit, in each surface form) and are used as input examples of
different contexts. The prompt explicitly asks for JSON format output and 20 samples, due to the sequence length
limitation of ChatGPT.  
The final prompt is as follows, where \texttt{UNIT1} and
\texttt{UNIT1} are replaced with different units with the shared
surface and \texttt{"SURFACE\_FORM"} denotes the ambiguous surface form:
\begin{verbatim}
Create a training set of 20 samples, for 
"UNIT1" and "UNIT2", where in the text the 
surface form of the unit is always 
"SURFACE_FORM", but the unit is different.
Output in JSON format as follows: 

{"text":"Sentence 1", "unit": "UNIT1" },
{"text":"Sentence 2 ", "unit": "UNIT2" }}
\end{verbatim}

\noindent   
The test dataset is created in the same manner.  For certain units,
multiple generations were required to get more complex sentences.  In
such cases, we specifically asked for sentences that do not start with
"the" and are more complex.  After each generation, all examples were
checked by the authors of the paper.  Faulty samples with wrong units
were removed. In some cases, surface forms were manually altered to
match the specification of the task.

\subsection{Disambiguation per Class}\label{sec:appendix5}

A detailed evaluation of the disambiguation dataset is shown in
Table~\ref{tab:perclass}, where precision, recall, and F1-score are
computed separately for each class.  For each surface form, 10
examples are present in the test dataset.  We noticed that
distinguishing between ``japanese yen'' and ``chinese yuan '' is
partially difficult for the BERT-based classifier since both of them
are currencies and used in similar contexts.  Another hard distinction
is between ``penny'' and ``point'', since monetary values and the
stock market point unit are used in similar contexts.  In comparison,
in Q3 certain units are almost never predicted, hence the multiple
zeros in the evaluation results.

\begin{table}[t]
\caption{Precision, recall and F1-score for \emph{unit} disambiguation per class.} 
\label{tab:perclass}
\setlength{\tabcolsep}{6pt} 
\renewcommand{\arraystretch}{1} 
\resizebox{0.5\textwidth}{!}{%
\begin{tabular}{lccc|ccc}
\toprule
 \multicolumn{1}{c}{\multirow{2}{*}{Class}} &\multicolumn{3}{c}{ CQE}& \multicolumn{3}{c}{Q3}  \\
 \cmidrule{2-7}
\multicolumn{1}{c}{}   & P&R&\multicolumn{1}{c}{F1} & P&R&F1   \\

\midrule
 knot & 100   &  100   &  100 &  16.67  &   100  &   28.57\\
roentgen  &   100   &  100  &   100 &  44.44   &  80  &   57.14\\
barn   &  100   &  80  & 88.89 &  100 &    80&     88.89\\
japanese yen  &    60  &   100   &  75 &100  &   100   &  100 \\
inch  &    83.33  &   100   &  90.91& 77.78  &   70  &   0.7368\\
armenian dram &     100 &    60  &   75& 0 &   0 &  0\\
chinese yuan  &    0   &  0   &  0 & 100 &    100  &   100\\
byte  &    100  &   100 &    100& 57.14  &   80  &   66.67\\
cent   &   83.33  &   100   &  90.91& 0 &  0 &  0\\
croatian kuna   &   100  &   100 &    100& 0  & 0  & 0\\
year    &  100   &  100  &   100& 0 &  0 &  0\\
poise   &   100  &   100&     100& 100 &   100  &  100\\
south african rand   &   100  &   100 &    100 & 100  &  60  &  75\\
minute &     80   &  80   &  80& 60 &   90  &  72\\
bit   &   83.33&     100    & 90.91 & 50  &  40  &  44.44\\
bel&      80   &  100    & 88.89& 100  &  100  &  100\\
kiloton    &  100  &   100  &   100& 100 &   100  &  100\\
second   &   100  &   80    &  88.89& 80 &   40  &  53.33\\
coulomb   &   100   &  100  &   100& 100  &  100  &  100\\
dram    &  71.43  &   100  &   83.33& 0  & 0 &  0\\
point&      55.56   &  100  &   71.43& 0 &  0  & 0\\
fahrenheit     & 100    & 100  &   100& 100    &100 &   100\\
celsius    &  100  &   90  &   94.74& 0  & 0   &0\\
pixel   &   100 &    100  &   100& 80  &  80   & 80\\
pound-mass   &   100  &   100 &    100& 83.33  &  100  &  90.91\\
pound sterling   &   100  &   100  &   100& 100   & 80  &  88.89\\
foot  &    80  &   80   &  80& 100  &  50   & 66.67 \\
penny   &   100  &   20    & 33.33& 0 &  0  & 0\\
farad    &  100 &    80    & 88.89& 80   & 80   & 80\\
acre    &  100  &   100   &   100& 0 &  0  & 0\\
\bottomrule
\end{tabular}%
}
\end{table}

\subsection{Error Analysis on NewsQuant and Txt-R
  Datasets}\label{sec:appendix6}

We analyzed the incorrect detection for all the models and the common
mistakes.  Except for the points discussed in the main part of the paper,
Table~\ref{tab:mistakes} provides an overview of the remaining common
mistakes and systems associated with them.

\end{document}